\documentclass[11pt]{article}
\usepackage{booktabs}

\usepackage{acl}
\usepackage{tikz}
\usepackage{stfloats}
\usepackage{fontawesome5}
\usetikzlibrary{positioning}

\usepackage{times}
\usepackage{latexsym}
\usepackage{amssymb}
\usepackage{pifont}
\usepackage{arydshln}

\usepackage[T1]{fontenc}

\usepackage[utf8]{inputenc}

\usepackage{microtype}

\usepackage{inconsolata}

\usepackage{graphicx}
\usepackage{xcolor}
\definecolor{darkgreen}{rgb}{0.0, 0.5, 0.0}

\usepackage{amsmath,amsfonts,bm}

\definecolor{mypink3}{cmyk}{0, 0.7808, 0.4429, 0.1412}

\definecolor{blue(pigment)}{rgb}{0.2, 0.2, 0.6}

\definecolor{crimsonglory}{rgb}{0.75, 0.0, 0.2}

\definecolor{britishracinggreen}{rgb}{0.0, 0.26, 0.15}

\definecolor{orange(colorwheel)}{rgb}{1.0, 0.5, 0.0}

\def\eqref#1{equation~\ref{#1}}

\def\1{\bm{1}}

\DeclareMathAlphabet{\mathsfit}{\encodingdefault}{\sfdefault}{m}{sl}
\SetMathAlphabet{\mathsfit}{bold}{\encodingdefault}{\sfdefault}{bx}{n}

\usepackage[dvipsnames]{xcolor}

\usepackage[most,skins,theorems]{tcolorbox}
\usepackage{enumitem}
\usepackage{amssymb}
\tcbset{
  aibox/.style={
    width=\linewidth,
    top=7pt,
    bottom=2pt,
    colback=blue!6!white,
    colframe=black,
    colbacktitle=black,
    enhanced,
    center,
    attach boxed title to top left={yshift=-0.1in,xshift=0.15in},
    boxed title style={boxrule=0pt,colframe=white,},
  }
}
\newtcolorbox{AIbox}[2][]{aibox,title=#2,#1}

\title{On the Role of Discreteness in Diffusion LLMs}

\author{
 \textbf{Ziqi Jin\textsuperscript{1,2}},
 \textbf{Bin Wang\textsuperscript{*1}},
 \textbf{Xiang Lin\textsuperscript{1}},
 \textbf{Lidong Bing\textsuperscript{1}},
 \textbf{Aixin Sun\textsuperscript{*2}}
\\
 \textsuperscript{1}MiroMind AI,
 \textsuperscript{2}Nanyang Technological University, Singapore
}

\begin{document}
\maketitle

\begin{figure*}[b]
  \centering
  \includegraphics[width=1.0\linewidth]{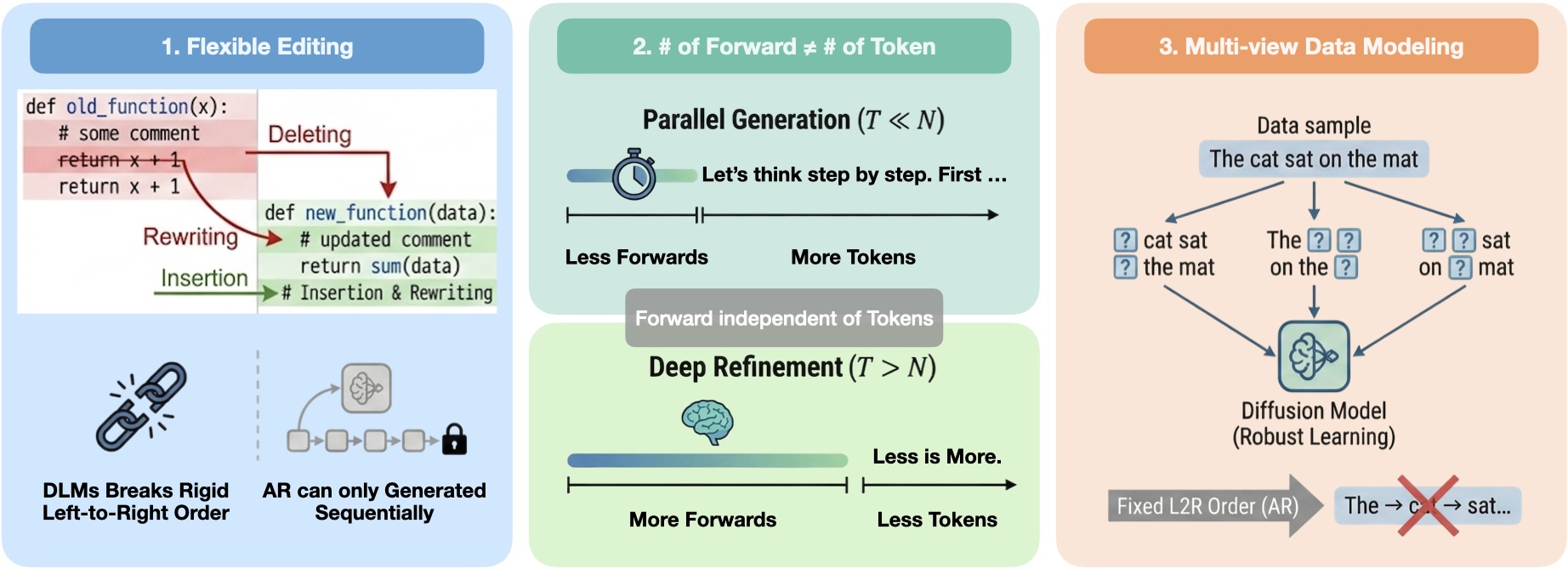}
  \caption{{\bf Core Advantages of Diffusion Language Models (DLMs) over Autoregressive (AR) Models.}}
  \label{fig:main_1}
\end{figure*}

\begin{abstract} 

Diffusion models offer appealing properties for language generation, such as parallel decoding and iterative refinement, but the discrete and highly structured nature of text challenges the direct application of diffusion principles. In this paper, we revisit diffusion language modeling from the view of diffusion process and language modeling, and outline five properties that separate diffusion mechanics from language-specific requirements. We first categorize existing approaches into \textit{continuous diffusion in embedding space} and \textit{discrete diffusion over tokens}. We then show that each satisfies only part of the five essential properties and therefore reflects a structural trade-off. Through analyses of recent large diffusion language models, we identify two central issues: (i) uniform corruption does not respect how information is distributed across positions, and  (ii) token-wise marginal training cannot capture multi-token dependencies during parallel decoding. These observations motivate diffusion processes that align more closely with the structure of text, and encourage future work toward more coherent diffusion language models.
\footnote{*Correspond: \href{mailto:bin.wang@miromind.ai}{bin.wang@miromind.ai}, \href{mailto:axsun@ntu.edu.sg}{axsun@ntu.edu.sg}}

\end{abstract}

\section{Introduction}
\label{sec:intro}

Language modeling is to learn a probability distribution over word sequences, $P(x_{1:n})$.  
Autoregressive (AR) models, which predict tokens from left to right, is currently trending.  
Recent work explores diffusion language models (DLMs) as an alternative modeling method. With diffusion process, DLMs generate text by reversing a noise-adding process. This raises a natural question: \textit{can diffusion be beneficial in language modeling?}

Diffusion offers potential advantages that AR models do not naturally support, illustrated in Figure~\ref{fig:main_1}.  
AR models generate strictly from left to right and add one token per forward pass.  In AR decoding, the model naturally extends a prefix instead of inserting, deleting, or rewriting earlier sequences. Diffusion, by updating many positions at once, can support \textit{flexible editing} such as insertion, deletion, and span rewriting more directly \cite{kim2025anyorderflexiblelengthmasked}.  AR decoding also ties computation directly to output length, since each new token needs one forward pass, whereas diffusion can revise multiple tokens together and adjust the number of refinement steps so that harder cases use more steps and easier ones use fewer.  Finally in the training stage, AR sees each token once in a fixed left-to-right order, which makes the task easy to fit but leads to faster overfitting in low-data settings; diffusion training instead shows each example at different noise levels and patterns due to randomness, creating multiple augmented variations of the same sequence, which improves data efficiency in the long run where data is limited \cite{ni2025diffusionlanguagemodelssuper,ni2025trainingoptimallargediffusion}.  We discuss these advantages in more detail in Section~\ref{sec:payoff}.

While the diffusion process can be applied naturally to visuals \cite{ho2020denoising}, applying diffusion to language introduces a core difficulty: text is discrete, but standard diffusion process assumes continuous data that can be corrupted with Gaussian noise smoothly.  
Because of this mismatch, existing DLMs only follow diffusion principles partially.  Specifically, 
\textit{continuous} DLMs  add Gaussian noise in embedding space, but the final step must map continuous vectors back to discrete tokens. This mapping is discontinuous and breaks the diffusion interpretation \cite{li2022diffusion, strudel2022self, dieleman2022continuous}.  
\textit{Discrete} DLMs avoid this issue by operating directly on tokens, but they replace continuous noise with masking \cite{austin2021structured, he2023diffusionbert}. Masking lacks the smooth, time-indexed corruption that diffusion relies on.

Most recent large-scale systems adopt masked discrete diffusion. While effective in practice, these models still leave open a basic question: \textit{what properties should an ideal diffusion language model have?}

In this paper, we revisit diffusion and language seperately.  
We identify five essential properties that a diffusion-based language model should satisfy in Section~\ref{sec:five_essentials}, separating general diffusion requirements from constraints that arise from discrete text.  
Through this lens, we show that both continuous and discrete DLMs satisfy only a subset of these requirements, and each involves trade-offs that lead to structural challenges for text diffusion. Our main contributions can be summarized as follows:

\begin{itemize}[leftmargin=1.2em, itemsep=0.1em]

    \item We propose a framework that clarifies what diffusion requires and what language imposes, and use it to reorganize existing diffusion language models into continuous and discrete families with a unified analysis of their assumptions and limitations.

    \item We provide theoretical and empirical evidence of structural mismatches between diffusion and text, including position-dependent difficulty under uniform corruption and the gap between marginal training and joint coherence.

    \item We outline several potential research directions related to these challenges, with the goal of informing future work on more complete and structurally aligned diffusion models for language.

\end{itemize}

\section{Preliminary}
\label{sec:pre}


Diffusion models \cite{ho2020denoising} define a generative process through two coupled components:
a \emph{forward corruption} that gradually destroys information, and a \emph{reverse model} that learns to recover data by iterative denoising.  
In the original continuous setting, the forward process injects small Gaussian noise so that a clean sample $x_0$ becomes increasingly noisy, and after $T$ steps the state approaches a simple prior (typically a standard Gaussian).  

A diffusion language model (DLM) applies the diffusion idea to text sequences $x_{1:n}$ over a vocabulary $\mathcal{V}$.  
The central design choice is the \emph{state space} where corruption and denoising happen.  
This leads to two main families: \textbf{continuous DLMs}, which diffuse in real-valued spaces, and \textbf{discrete DLMs}, which diffuse directly over tokens.  
They differ primarily in (i) how noise is defined, and (ii) what the reverse model predicts at each step.

\paragraph{Continuous DLMs}
represent text as a real-valued sequence, such as token embeddings or other continuous vectors, and apply Gaussian diffusion to these representations
\cite{li2022diffusion, strudel2022self, dieleman2022continuous}.  
A typical workflow is:

\begin{itemize}[leftmargin=1.2em, itemsep=0.1em]
\item \textit{State:}  a continuous sequence $z_0$ derived from text (e.g., word embeddings, one-hot embeddings, or other continuous encodings).
\item \textit{Forward:} add Gaussian noise to obtain $z_t$ at different noise levels.
\item \textit{Training:} learn a denoiser that predicts a clean target (often $z_0$ or an equivalent parameterization) from $z_t$.
\item \textit{Generation:} start from Gaussian noise $z_T$ and iteratively denoise to $z_0$, then convert the final continuous state into tokens.
\end{itemize}

The appeal is that continuous DLMs inherit the original diffusion structure: smooth corruption and joint refinement over all positions.  
At the same time, language information has a discrete component (token identity) that interacts with Gaussian noise differently from continuous geometry.  
CANDI \cite{pynadath2025candi} highlights this tension: the signal that preserves token identity and the signal that supports smooth continuous denoising can become useful at different noise ranges, making it difficult for a single Gaussian diffusion trajectory to serve both roles well.

\paragraph{Discrete DLMs} keep the state in the token domain and define corruption using masking or categorical transition kernels
\cite{austin2021structured, he2023diffusionbert}.  
A common instance is masked discrete diffusion: at higher noise levels, more positions are replaced by a special mask token.  
The workflow is:

\begin{itemize}[leftmargin=1.2em, itemsep=0.1em]
\item \textit{State:} a token sequence $x_t \in \mathcal{V}^n$.
\item \textit{Forward:} increase uncertainty by replacing tokens with a mask (or by sampling from a categorical transition).
\item \textit{Training:} learn a denoiser that predicts token distributions for corrupted positions given the partially observed sequence.
\item \textit{Generation:} begin from a highly corrupted sequence (e.g., mostly masks) and iteratively fill in / refine tokens over multiple steps.
\end{itemize}

Discrete DLMs align naturally with the discreteness of language, but their corruption is inherently \emph{stepwise}: a token is kept, replaced, or transitioned among categories, rather than being perturbed infinitesimally as in Gaussian diffusion.  
Recent work scales masked discrete diffusion to large models \cite{nie2025large, ye2025dream}, and explores improved training objectives and discrete-time formulations that better fit categorical state spaces \cite{lou2024discrete, sahoo2024simple, gat2024discrete}.  
Finally, hybrid systems inject diffusion-style updates into autoregressive decoding to combine parallel refinement with sequential structure
\cite{han2023ssd, wu2023ar}.

\section{Advantages of DLMs}
\label{sec:payoff}

Diffusion language models offer several advantages that are difficult to realize with autoregressive (AR) models.  
These benefits stem from the fact that diffusion updates all positions jointly rather than committing to tokens one at a time.  
We highlight three central advantages: flexible editing, decoupled compute–length scaling, and improved data efficiency.

\paragraph{Flexible Editing.}
AR models generate text strictly left-to-right; once a token is produced, it cannot be revised without regenerating the entire sequence. This constraint prevents global adjustments and limits interactive editing.  
Diffusion models, by contrast, treat the entire sequence as mutable throughout the denoising trajectory.  
Each step jointly refines all positions, enabling natural support for infilling, rewriting, and any-order generation.

This property is particularly valuable for more structured text domains such as programming codes.  
A modification to a function signature or variable declaration may require coordinated changes across multiple distant locations.  
Diffusion's joint refinement allows such global consistency updates without restarting generation, offering a practical advantage over causal models that can only append tokens.

\paragraph{Token generation is not one-to-one with forward operations.} 
AR inference requires one forward pass per token, coupling computation to sequence length and preventing dynamic allocation of compute based on task difficulty.  
Diffusion breaks this constraint by introducing a refinement process with $T$ steps independent of the output length $N$, which includes two aspects:

\begin{itemize}[leftmargin=1.2em, itemsep=0.2em]
    \item \textit{Parallel generation ($T \ll N$):}  
    A long sequence can be generated simultaneously, providing substantial speedups for long-context synthesis.
    \item \textit{Deep refinement with more generation steps than the number of tokens ($T > N$):}  
    For tasks with high information density such as multi-step reasoning, long-form planning, or drafting structured reports, the model can spend more refinement steps per token.  
    This matches the intuition that harder problems demand more ``thinking time,'' aligning with test-time scaling practices without excessively increasing the number of tokens.
\end{itemize}

\paragraph{Multi-View Data Modeling.}
Recent work shows that DLMs can use data more efficiently than
autoregressive models, especially when the training set is small
\cite{ni2025diffusionlanguagemodelssuper, ni2025trainingoptimallargediffusion}.  
This arises not only because noise acts as data augmentation, but also because diffusion does not constrain the model to a single left-to-right factorization of the sequence.

If we train causal transformers with different token orders, left-to-right (L2R)
and right-to-left (R2L) models learn at almost the same speed, while random
orders learn much more slowly \cite{cunxiao}.  
This suggests that L2R is not a fundamental property of language, but a modeling
choice that happens to work well because it matches local word-to-word
dependencies.
Using a fixed order (L2R or R2L) makes the prediction task very regular:
every example is always seen as a prefix followed by a target, which makes it
easy for the model to fit the training data, but also makes it easier to
memorize.

Diffusion breaks this fixed ordering with random masking.
Each sequence is seen many times with different masked positions, so the model
must recover tokens from many kinds of partial context instead of a single,
fixed prefix.
This acts as a strong form of data augmentation.
In practice, autoregressive models usually reach good performance faster but
start to overfit earlier, while diffusion models improve more slowly and keep
benefiting from additional epochs, which matches the idea that their supervision
signal is broader and more challenging.

\begin{table*}
\centering
\begin{tabular}{p{4.5cm}lll}
\toprule
\textbf{Criterion} & \textbf{AR} & \textbf{Continuous DLMs} & \textbf{Discrete DLMs} \\
\midrule
\multicolumn{4}{l}{\textit{\textbf{\ Diffusion Properties (D)}}} \\
(\textbf{D1}) Smooth Corruption 
& \textcolor{red}{\ding{55}} (Not diffusion) 
& \textcolor{darkgreen}{\ding{51}} (Infinitesimal) 
& \textcolor{red}{\ding{55}} (Stepwise/Coarse) \\
(\textbf{D2}) Tractable Intermediate States 
& \textcolor{darkgreen}{\ding{51}} (Prefix state) 
& \textcolor{darkgreen}{\ding{51}} (Gaussian) 
& \textcolor{darkgreen}{\ding{51}} (Categorical) \\
(\textbf{D3}) Iterative Refinement 
& \textcolor{darkgreen}{\ding{51}} (Token-by-token) 
& \textcolor{darkgreen}{\ding{51}} (Latent trajectory) 
& \textcolor{darkgreen}{\ding{51}} (Token trajectory) \\
\addlinespace[0.5em]
\hdashline
\multicolumn{4}{l}{\textit{\textbf{\ Language Properties (L)}}} \\
(\textbf{L1}) Discreteness 
& \textcolor{darkgreen}{\ding{51}} (Tokens) 
& \textcolor{red}{\ding{55}} (Continuous states) 
& \textcolor{darkgreen}{\ding{51}} (Tokens) \\
(\textbf{L2}) Structural Dependency 
& \textcolor{orange}{\ding{51}}\textsuperscript{\textcolor{orange}{\kern-0.5em\tiny\ding{55}}} (Casual) 
& \textcolor{orange}{\ding{51}}\textsuperscript{\textcolor{orange}{\kern-0.5em\tiny\ding{55}}} (Implicit) 
& \textcolor{orange}{\ding{51}}\textsuperscript{\textcolor{orange}{\kern-0.5em\tiny\ding{55}}} (Implicit) \\
\bottomrule
\end{tabular}
\caption{AR and diffusion language models (continuous and discrete) viewed through \textbf{Diffusion Properties} and \textbf{Language Properties}. AR satisfies language properties but does not define a diffusion-style smooth corruption path. Continuous DLMs preserve smooth diffusion but operate in continuous states; discrete DLMs remain token-based but use stepwise corruption and marginal denoising.}
\label{tab:true_diffusion_essentials}
\end{table*}

\section{Properties for Diffusion LLMs}
\label{sec:five_essentials}

Diffusion models are defined by a tight link between how data are \emph{corrupted}
and how they are \emph{recovered}. When we move from images to text, the same
idea remains attractive, but it must be reconciled with what diffusion assumes
and what language is. To keep the discussion concrete, we summarize five
properties that are used throughout the paper: three describe the diffusion
mechanism itself, and two come from the nature of text.

In continuous domains, diffusion uses a forward noising process and a learned
reverse denoising process. The forward process adds small noise to gradually
increase uncertainty, and intermediate noisy states can be sampled at arbitrary
noise levels without simulating the full chain. The reverse process starts from
a simple noise prior and iteratively denoises to recover a clean sample.
We refer to these diffusion-side properties as \textbf{D1: smooth corruption},
meaning the time index corresponds to gradual, continuous noise changes rather
than abrupt jumps; \textbf{D2: tractable intermediate states}, meaning the
marginal corruption distribution $q(x_t \mid x_0)$ is available in closed form or
through an analytic procedure so training can sample $x_t$ directly; and
\textbf{D3: iterative reverse generation}, meaning generation starts from
a simple noise prior and repeatedly applies learned reverse updates to refine the
same state over multiple steps. Text, in contrast, introduces two properties
that are independent of any modeling choice: \textbf{L1: discreteness},
since text is composed of discrete symbols and changing a token is a jump rather
than an infinitesimal perturbation, and \textbf{L2: structural dependency},
since syntax and semantics impose long-range constraints that couple positions.

\subsection{How Current DLMs Fit the Properties}

Table~\ref{tab:true_diffusion_essentials} summarizes how autoregressive (AR)
models and diffusion language models (DLMs) align with the diffusion-side
properties (D1--D3) and the language-side properties (L1--L2).

AR models generate sequences by a sequential factorization, predicting each token
conditioned on its prefix. They therefore satisfy  {discreteness (L1)} by
operating directly on tokens. For  {structural dependency (L2)}, AR imposes
an explicit \emph{causal} dependency assumption: it hard-codes one specific
factorization of dependencies (left-to-right), which captures an important class
of linguistic constraints, but does not represent the full space of global
dependencies in a symmetric way. On the diffusion side, AR is not diffusion in
the sense of  {smooth corruption (D1)}, but it does have a simple and
tractable notion of intermediate state: the current prefix together with the
un-generated suffix ( {D2}). Generation is also iterative by construction
( {D3}), refining the sequence token by token.

Continuous DLMs apply diffusion to continuous representations derived from text,
such as embeddings, latent vectors, or token-distribution states
\cite{ho2020denoising,li2022diffusion,strudel2022self,dieleman2022continuous,gongdiffuseq}.
They closely follow the original diffusion mechanism: Gaussian-style
perturbations yield  {smooth corruption (D1)}; intermediate marginals are
 {tractable (D2)} in the continuous state space; and sampling proceeds
through  {iterative refinement (D3)} along a latent denoising trajectory.
On the language side, however, the diffusion state is continuous rather than
symbolic, so  {discreteness (L1)} is not satisfied. For  {structural
dependency (L2)}, these models typically do not impose an explicit dependency
factorization; instead, any multi-token constraints must be learned implicitly
through the denoising network.

Discrete DLMs define diffusion directly over tokens using masking or categorical
transition kernels \cite{austin2021structured,he2023diffusionbert}. They satisfy
 {discreteness (L1)} by construction. On the diffusion side, because the
corruption process is specified by explicit transition matrices, the intermediate
corruption distribution is  {tractable (D2)}, and generation follows
 {iterative refinement (D3)} with token-level intermediate states. However,
 {smooth corruption (D1)} is  approximated: even with continuous-time
parameterizations \cite{austin2021structured,lou2024discrete}, the underlying
state changes remain stepwise due to the discrete space, and recent analysis
suggests that $t$ often behaves like a proxy for the number of masked tokens
rather than a smooth signal-to-noise ratio \cite{zheng2024masked}. For
 {structural dependency (L2)}, discrete DLMs also rely mainly on implicit
learning rather than an explicit dependency assumption: the corruption is usually
position-symmetric, and dependency structure is expected to emerge from the
denoiser rather than from the generative factorization itself.

\paragraph{Trade-offs.} Viewed through these properties, continuous and discrete DLMs reflect a recurring
trade-off. Continuous methods stay close to classical diffusion and preserve
(D1 -- D3), but they move away from symbolic text (L1) and usually leave (L2) to
be learned implicitly. Discrete methods stay faithful to symbols (L1) and keep
tractable corruption (D2), but they sacrifice smoothness (D1) and do not
build in a dependency structure (L2). Soft-Masked Diffusion
\cite{hersche2025softmaskeddiffusionlanguagemodels} illustrates how improving one
property can weaken another: by replacing binary masks with soft mixtures over
candidate tokens, it makes the corruption trajectory more gradual and improves
(D1), but the intermediate corruption distribution becomes model-dependent and is
no longer available in closed form, weakening (D2) and requiring a two-pass
training scheme. Overall, we conclude that property-level
Trade-offs should be explicitly considered when designing methods that incorporate both language and diffusion.

\begin{figure*}[ht!]
  \centering
  \includegraphics[width=1.0\linewidth]{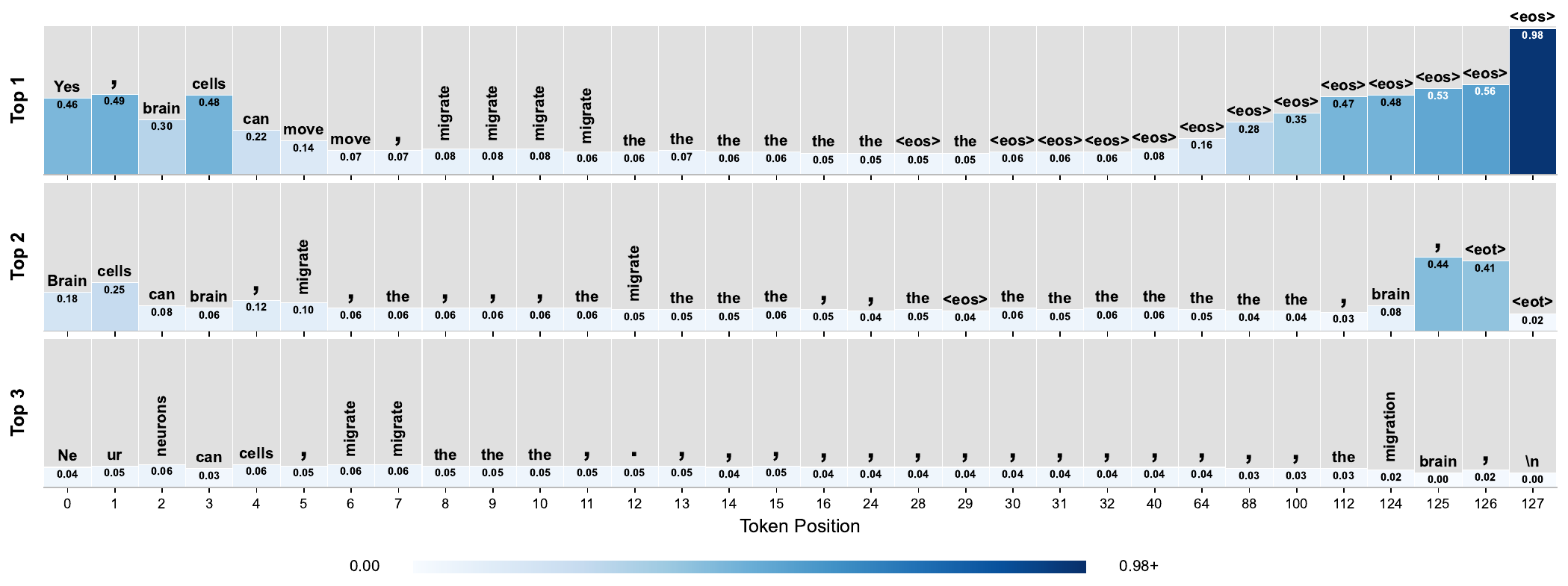}
  \caption{An example from LIMA dataset \cite{zhou2023lima} illustrating the probability distribution analysis of top-3 tokens by LLaDA-Instruct with 128 masked positions when prompt with "\textit{\textbf{Can brain cells move? By movement I mean long distance migration (preferably within the brain only).}}". We found early {\tt [MASK]} are more determistic, while distant ones collapse to frequency-dominant. The Experiment Setting is detailed in Appendix \ref{sec:exp_detail}.}
  \label{fig:tok_analysis}
\end{figure*}

\section{Core Challenges for DLMs}

The previous sections motivate diffusion for language and summarize the key
properties in Table~\ref{tab:true_diffusion_essentials}.  We now focus on the
two properties that are most problematic in practice and largely determine the
current gap between diffusion and text: \textbf{smooth corruption} (D1) and
\textbf{structural dependency} (L2).  We first discuss why designing a
``smooth'' corruption process for discrete sequences is non-trivial
(Section~\ref{sec:deep_dive_1}), and then analyze why token-wise, parallel
denoising struggles to capture multi-token constraints (Section~\ref{sec:deep_dive_2}).



\subsection{Smooth Corruption does not mean Even Information Loss}
\label{sec:deep_dive_1}

In Section~\ref{sec:five_essentials}, we defined (D1) \emph{Smooth Corruption} from a modeling perspective.  
Here we view it from an information perspective: a good noise schedule should make the information about the target decay gradually as the noise level $t$ increases.  
In other words, when $t$ changes a little, the amount of recoverable information should also change a little, both at the sequence level and at the token level.

For language, this is challenging because information is not evenly distributed across tokens.
Some tokens carry most of the meaning and strongly constrain the rest of the sentence, while others are easier to infer.
This is also why attention matters: it lets the model weigh context by usefulness instead of treating all positions equally.
If corruption applies the same rule everywhere, then ``the same noise level'' does not mean ``the same information loss''.
Corrupting a few high-impact tokens can destroy the core meaning early, while corrupting low-impact tokens may change little.
We examine this mismatch at both the sequence level and the token level.

At the sequence level, discrete and continuous diffusion behave quite differently. In masked discrete diffusion, the forward process gradually increases the number of masked positions.  
Even at high mask ratios, a subset of tokens may remain visible, so some coarse information about the sequence is preserved.  
However, this preservation is blind to where information actually lies: all positions are equally likely to be masked, regardless of whether they carry crucial content or redundant tokens.  
The same noise level $t$ can therefore correspond to very different amounts of remaining information, depending on which positions were masked. Continuous diffusion approaches apply Gaussian noise to every coordinate in an embedding or one-hot space.  
From a mathematical standpoint this is perfectly smooth, but CANDI \cite{pynadath2025candi} shows that, for large vocabularies, the discrete identity of a token becomes unrecoverable at relatively low noise levels.  
The model’s ability to distinguish the correct token from tens of thousands of alternative tokens collapses quickly, even though the continuous signal still changes slowly.  
Thus, if we measure smoothness in terms of \emph{recoverable information} rather than variance of the noise, the corruption is not gradual: token identity is lost in a few steps.

At the token level, the situation is more uneven.
In discrete diffusion, corruption is often described as ``binary'' (keep or mask).  
However, from an information viewpoint, a masked token is not always equally unknown.  
If its neighbors are  visible, local context strongly constrains what the token can be; if its neighbors are also masked, the same token effectively experiences a much higher noise level.  
Uniform masking therefore produces a wide spread of effective noise across positions: some tokens remain easy to recover, while others quickly become almost impossible.

This explains the ``frequency collapse'' observed in practice.  
Let $x_i$ be a token and $x_{\mathcal{O}}$ the set of positions that survive corruption.  
A diffusion LM estimates $p(x_i \mid x_{\mathcal{O}})$.  
As more local context is removed around $x_i$, the Mutual Information $I(x_i; x_{\mathcal{O}})$ rapidly decreases.  
When the remaining context carries almost no information about $x_i$, we have
\begin{equation}
    \lim_{I(x_i; x_{\mathcal{O}}) \to 0} p(x_i \mid x_{\mathcal{O}}) = p(x_i),
\end{equation}
so the optimal prediction becomes the marginal distribution.  
As a result, the model prefers very common tokens (such as ``the'', punctuation, or \texttt{<eos>}) because, given the corrupted input, these are simply the statistically safest guesses.

\paragraph{Empirical Study.}

To illustrate such token-level locality, our analysis of how current DLMs treats masked token makes this effect visible  
(Figure~\ref{fig:tok_analysis}).  
We prompt the model with  
\emph{``Can brain cells move? By movement I mean long distance migration (preferably within the brain only).''}  
and then append 128 masked positions.  
We inspect the top-3 predictions at each position.

At positions closest to the prompt (0--2), the predictions are confident and semantically appropriate, such as ``Yes'' and ``brain''.  
Here, nearby context provides strong constraints, so the model can effectively recover the token.  
As we move further away (positions 12--29), the influence of the prompt fades.  
The predictions become uncertain and collapse toward high-frequency tokens such as ``the'' and punctuation, even though the semantic question has not changed.  
At even more distant positions, the model assigns high probability to \texttt{<eos>}, reflecting only the dataset’s length statistics.  
A similar pattern appears in other DLMs such as Dream-7B: early positions are tightly constrained, while distant ones drift toward the unigram prior.

This experiment highlights that, under a uniform corruption schedule, local information disappears much faster than the nominal noise level would suggest.  
Positions that are equally ``noised'' according to $t$ can have very different amounts of usable information, depending on how much local context remains.


We found that some current methods have tried to mitigate such effect. For example, Dream-7B explicitly addresses this mismatch \cite{ye2025dream}.  
Its Context-Adaptive noise Rescheduling at Token-level (CART) scales the training loss by a geometric function of the distance to the nearest unmasked token.  
Positions with nearby anchors are upweighted, encouraging the model to learn from contexts where recovery is feasible,  
while positions surrounded by masks are downweighted, preventing the model from overfitting to signals that are effectively just word frequency.  
In our words, CART modifies the training objective to not over-punish the mask with little context.

\paragraph{Future Direction.}

These observations suggest that satisfying (D1) for language requires more than a continuous noise schedule;  
the corruption must be smooth in terms of information, not only in terms of variance.

For discrete DLMs, one direction is to move beyond binary masking and define transition kernels that change tokens in smaller, structured steps,  
for example along semantic or categorical axes (specific $\to$ general $\to$ \texttt{[MASK]}).  
Such processes could keep token identity recoverable for longer while still increasing noise.

For continuous DLMs, hybrids such as CANDI \cite{pynadath2025candi} decouple discrete identity from continuous refinement.  
They maintain a corruption process that preserves which token is present, while a separate continuous channel learns smooth gradients.  
Both lines of work share the same goal: to align the corruption process with how information is distributed across positions,  
so that information loss becomes more gradual and even, at both the sequence and token level.

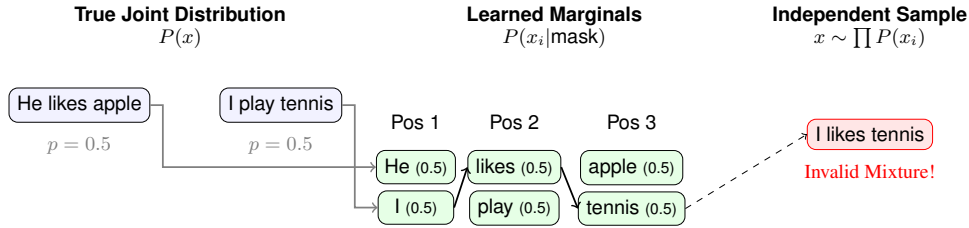
\begin{figure*}[t]
\centering
\resizebox{0.8\linewidth}{!}{%
\begin{tikzpicture}[
    node distance=1.5cm and 2cm,
    font=\small\sffamily,
    token/.style={draw, rounded corners, fill=blue!5, align=center, minimum width=1.2cm},
    prob/.style={text=gray, font=\footnotesize},
    cross/.style={cross out, draw=red, minimum size=2*(#1-\pgflinewidth), inner sep=0pt, outer sep=0pt},
    cross/.default={6pt}
]

\node[align=center] (true) {\textbf{True Joint Distribution}\\$P(x)$};
\node[token, below=0.5cm of true, xshift=-1.6cm] (seq1) {He likes apple};
\node[token, below=0.5cm of true, xshift=1.6cm] (seq2) {I play tennis};
\node[below=0.1cm of seq1, prob] {$p=0.5$};
\node[below=0.1cm of seq2, prob] {$p=0.5$};

\node[align=center, right=2.6cm of true] (marginal) {\textbf{Learned Marginals}\\$P(x_i | \text{mask})$};

\node[below=0.8cm of marginal, xshift=-2.2cm] (pos1_label) {Pos 1};
\node[token, below=0.2cm of pos1_label, fill=green!10] (he) {He \scriptsize{(0.5)}};
\node[token, below=0.1cm of he, fill=green!10] (i) {I \scriptsize{(0.5)}};

\node[right=0.5cm of pos1_label] (pos2_label) {Pos 2};
\node[token, below=0.2cm of pos2_label, fill=green!10] (likes) {likes \scriptsize{(0.5)}};
\node[token, below=0.1cm of likes, fill=green!10] (go) {play \scriptsize{(0.5)}};

\node[right=0.8cm of pos2_label] (pos3_label) {Pos 3};
\node[token, below=0.2cm of pos3_label, fill=green!10] (apple) {apple \scriptsize{(0.5)}};
\node[token, below=0.1cm of apple, fill=green!10] (farm) {tennis \scriptsize{(0.5)}};

\node[align=center, right=1.8cm of marginal] (output) {\textbf{Independent Sample}\\$x \sim \prod P(x_i)$};
\node[token, below=1.0cm of output, fill=red!10, draw=red] (bad) {I likes tennis};
\node[below=0.1cm of bad, text=red, font=\footnotesize] {Invalid Mixture!};

\draw[->, thick, gray] (seq1.east) -- ++(0.2,0) |- (he.west);

\draw[->, thick, gray] (seq2.east) -- ++(0.2,0) |- (i.west);

\draw[->, thick] (i.east) -- (likes.west);
\draw[->, thick] (likes.east) -- (farm.west);
\draw[->, dashed] (farm.east) -- (bad.west);

\end{tikzpicture}
}
\caption{The ``{\bf Marginal Trap}'': A toy example shows that the model learns from ``He likes apple'' and ``I play tennis'' (50\% each). However, parallel decoding samples them independently resulting each position to be 50\% for both samples at each tokon position. Directly sampling form these distribution may create a path (I $\to$ likes $\to$ tennis) that never existed in the training data.}
\vspace{-1em}
\label{fig:marginal_trap}
\end{figure*}

\subsection{Absence of Structural Dependency}
\label{sec:deep_dive_2}

Masked discrete diffusion models typically learn token-wise conditionals given the visible context.
At a denoising step, the model outputs a separate distribution for each masked position,
$p(x_i \mid x_{\mathcal{O}})$, and the training loss is a sum of per-token cross-entropies.
Under this objective, the model is not directly trained to represent how multiple unknown tokens
should constrain one another.
As a result, the model can match the \emph{marginal} distribution at each position while still
missing \emph{joint} constraints required by language (L2), such as agreement and phrase-level
compatibility.

\paragraph{Conditions. }
Importantly, this limitation is most visible under two practical choices that are common in current masked DLMs.

\emph{(1) Committed intermediate states.}
Many implementations represent intermediate states as \emph{partially filled token sequences}:
once a token is sampled (or greedily chosen) at some step, it becomes part of the visible context
for later steps.
This commitment makes the model sensitive to early inconsistent choices, because later updates
must condition on them rather than revising them jointly.
By contrast, an idealized refinement process that keeps states \emph{uncommitted} until
convergence (e.g., operating on soft token distributions instead of hard tokens) does not force
such early irreversible decisions, and therefore reduces the chance of forming incompatible
multi-token combinations.

\emph{(2) Parallel updates with fewer steps than tokens.}
If the model updates many masked positions in parallel and uses a small number of denoising steps
($T \ll N$), then multiple dependent tokens must be decided at the same time, without an external
factorization that enforces their compatibility.
In contrast, if decoding updates only \emph{one token at a time} (or uses $T \ge N$ with a
sequential schedule), then joint constraints can be enforced implicitly through conditioning:
later choices depend on earlier chosen tokens, reducing inconsistent mixtures.
This resembles the role played by sequential factorization in AR decoding.

\paragraph{Example of the ``Marginal Trap.''}
Figure~\ref{fig:marginal_trap} shows a toy dataset containing only two sentences:
\emph{``He likes apple''} and \emph{``I play tennis''}.
The optimal token-wise model learns the correct marginals.
Each prediction is correct on its own, but sampling these marginals independently can produce
invalid mixtures like \emph{``I likes tennis''}.
This illustrates the core gap: marginals are correct, but the joint distribution is not modeled.

\paragraph{Empirical Evidence.}
The same pattern appears in large masked DLMs.
In LLaDA-8B-Instruct (Figure~\ref{fig:tok_analysis}), two nearby masked positions may both assign
non-trivial probability to the same token (e.g., \emph{``brain''}), so parallel sampling can yield
local duplications such as \emph{``brain brain''}.
This is not because each position is unreasonable in isolation; it is because the model has no
explicit mechanism to couple the decisions across positions during a parallel update.

In practice, large masked DLMs often rely on generating one token per forward or extremely low temperatures
\cite{nie2025large,ye2025dream} to reduce these inconsistent combinations.
Alternatively, one can reduce the issue by decoding with more sequential schedules (e.g., updating
one token per step), but this partially sacrifices the parallelism that motivates diffusion decoding.

\paragraph{Future Direction.}
Addressing (L2) requires mechanisms that couple multiple positions beyond token-wise losses. One direction is to move from per-token cross-entropy toward sequence-level or structured objectives that score joint configurations, so inconsistent multi-token outcomes are directly penalized (e.g., energy-based or contrastive formulations \cite{yang2021universalsentencerepresentationlearning}). A complementary direction is to use state representations that delay commitment, such as keeping intermediate states as soft token distributions, so later steps can revise earlier choices and correct inconsistencies jointly instead of being locked into early decisions. Finally, parallel decoding can be made more reliable by explicitly pushing the model toward convergence of single path. For example, dParallel trains a dLLM with a certainty-forcing distillation objective, encouraging many masked positions to reach high confidence earlier \cite{chen2025dparallel}.

\section{Conclusion}
\label{sec:conclusion}

In this paper, we revisited diffusion language models through a unified lens based on 
\textbf{Diffusion Properties (D1 to D3)} and \textbf{Language Properties (L1 and L2)}.
This framework shows that current diffusion LLMs occupy only parts of the ideal design landscape.  
Continuous approaches satisfy the mathematical form of diffusion but lose contact with the discrete and dependency-rich structure of text.  
Discrete approaches preserve the state space of language but must approximate diffusion through coarse masking and independent token predictions.

Our empirical analysis demonstrates that these structural gaps have direct impact during inference.  
In Section~\ref{sec:deep_dive_1}, we documented the phenomenon of frequency collapse:  
uniform corruption ignores how information is distributed across positions, causing recoverability to drop abruptly and pushing predictions toward the unigram prior.  
In Section~\ref{sec:deep_dive_2}, we highlighted the marginal trap:  
training objectives that operate on individual tokens cannot enforce multi-token dependencies, 
which leads parallel decoding to produce sequences that are locally plausible but globally inconsistent.  
Together, these findings show that the typical diffusion assumptions of uniform corruption and marginal denoising are not naturally aligned with the structure of text.

\section*{Limitations}

Our analysis is primarily conceptual and aims to clarify definitions, assumptions, and trade-offs
in diffusion language modeling. As a result, several limitations should be noted.
First, the ``five properties'' is an abstraction that compresses a wide range of diffusion
formulations into a small set of criteria; different choices of definitions may yield alternative interpretation. Second, our discussion reflects common design patterns in existing continuous and
discrete DLMs, but does not exhaust all variants (e.g., alternative state spaces, transition kernels,
or decoding schedules). Third, we did not take all variants of DLMs into consideration and discuss in detail due to the number of variation. Finally, we do not quantify end-to-end trade-offs among speed, quality, and controllability
under unified implementations, which would be required for strong engineering recommendations.

\bibliography{custom}

\appendix

\section{Experiment Details}
\label{sec:exp_detail}

To produce Figure~\ref{fig:tok_analysis}, we probe a masked DLM with a single forward pass under a fully-masked answer span. Concretely, we first format the user question using the model’s chat template, and then append $N=128$ mask tokens to represent an unknown response. The resulting input has the form
\[
\texttt{input} \;=\; \texttt{chat\_template(user)} \;\Vert\; [\texttt{MASK}]^{128}.
\]
We run the model once and extract the output distribution at each masked position. For every position $i \in \{0,\dots,127\}$, we apply softmax to the logits and record the top-3 tokens and their probabilities. Figure~\ref{fig:tok_analysis} visualizes these top-3 predictions across positions for one representative prompt.

Beyond the single example shown, we repeat the same procedure on 100 different prompts from LIMA's training dataset \cite{zhou2023lima}, and observe the same qualitative pattern: early masked positions tend to have sharper, more content-specific distributions, while distant positions become increasingly dominated by high-frequency tokens and special symbols such as punctuation or \texttt{<eos>}.

\end{document}